\documentclass{article}

     \PassOptionsToPackage{numbers, compress}{natbib}

     \usepackage[preprint]{neurips_2020}

\usepackage[utf8]{inputenc} 
\usepackage[T1]{fontenc}    
\usepackage{hyperref}       
\usepackage{url}            
\usepackage{booktabs}       
\usepackage{amsfonts}       
\usepackage{nicefrac}       
\usepackage{microtype}      
\usepackage{graphicx}
\usepackage[rightcaption]{sidecap}

\newcommand{\beginsupplement}{
        \setcounter{table}{0}
        \renewcommand{\thetable}{S\arabic{table}}
        \setcounter{figure}{0}
        \renewcommand{\thefigure}{S\arabic{figure}}
        \setcounter{section}{0}
        \renewcommand{\thesection}{\Alph{section}}
     }
     
\title{Graph Neural Networks Including Sparse Interpretability}

\author{%
  Chris ~Lin\\
  AstraZeneca\\
  \texttt{chris.lin@astrazeneca.com}\\
  \And
  Gerald J.~Sun\\
  AstraZeneca\\
  \texttt{gerald.sun@astrazeneca.com}
  \And
  Krishna C.~Bulusu\\
  AstraZeneca\\
  \texttt{krishna.bulusu@astrazeneca.com}
  \And
  Jonathan R.~Dry\\
  Tempus\\
  \texttt{dryscilab@gmail.com}
  \And
  Marylens ~Hernandez\\
  AstraZeneca\\\
  \texttt{marylens.hernandez@astrazeneca.com}
}

\begin{document}

\maketitle

\begin{abstract}
    Graph Neural Networks (GNNs) are versatile, powerful machine learning methods that enable graph structure and feature representation learning, and have applications across many domains. For applications critically requiring interpretation, attention-based GNNs have been leveraged. However, these approaches either rely on specific model architectures or lack a joint consideration of graph structure and node features in their interpretation. Here we present a model-agnostic framework for interpreting important graph structure and node features, \underline{G}raph neural networks \underline{I}ncluding \underline{S}par\underline{S}e in\underline{T}erpretability (GISST). With any GNN model, GISST combines an attention mechanism and sparsity regularization to yield an important subgraph and node feature subset related to any graph-based task. Through a single self-attention layer, a GISST model learns an importance probability for each node feature and edge in the input graph. By including these importance probabilities in the model loss function, the probabilities are optimized end-to-end and tied to the task-specific performance. Furthermore, GISST sparsifies these importance probabilities with entropy and L1 regularization to reduce noise in the input graph topology and node features. Our GISST models achieve superior node feature and edge explanation precision in synthetic datasets, as compared to alternative interpretation approaches. Moreover, our GISST models are able to identify important graph structure in real-world datasets. We demonstrate in theory that edge feature importance and multiple edge types can be considered by incorporating them into the GISST edge probability computation. By jointly accounting for topology, node features, and edge features, GISST inherently provides simple and relevant interpretations for any GNN models and tasks.
\end{abstract}

\section{Introduction}

Graphs are universal data structures that can capture almost any type of information or relationships among individual entities of a given dataset. Graphs can be used to represent social relationships, molecular structures, and biological interactions; their application to specific problems spans multiple fields from computer science, to chemistry, to biology, and to drug development \citep{Ying_2018, Kovacs2019, Zitnik_2018, Goh8685, gysi2020network, hamilton2017representation, you2018graph}. Despite their ability to efficiently store different entities, features, and relationships at multiple complexity scales, graphs could benefit from analytical tools that interrogate or model their data in useful, meaningful ways.

As such, computation over graph-structured data has become an area of active research. Resulting graph modeling frameworks have led to applications such as recommendation engines and marketing, or discovery of novel drugs targets and disease biomarkers. One recent graph modeling framework, Graph Neural Networks (GNNs), has achieved state-of-the-art performance by learning recursively over both the relational graph structure (i.e. neighboring nodes and their connections) and associated features \citep{xu2018powerful, battaglia2018relational, Goyal_2018}. 

However, GNN-based methods are unable to explain the underlying graph structure that contributes to model output predictions, thus lacking the transparency required for informed decision making. Current methods for GNN model interpretation are generally post-hoc, rely on specific model architectures, or lack a joint consideration of graph structure and node features.

\begin{SCfigure}[0.5][h]
    \centering
    \caption{GISST overview schematic. Sparsification of input graph's features and edges is achieved through the additions to the model loss function; this enables application of GISST to any graph neural network model and learning task. During model training, this also allows for coupling of model performance with generation of sub-featureset and sub-graph importance probabilities, and thus permits inherent interpretability.}
    \includegraphics[width=0.6\textwidth]{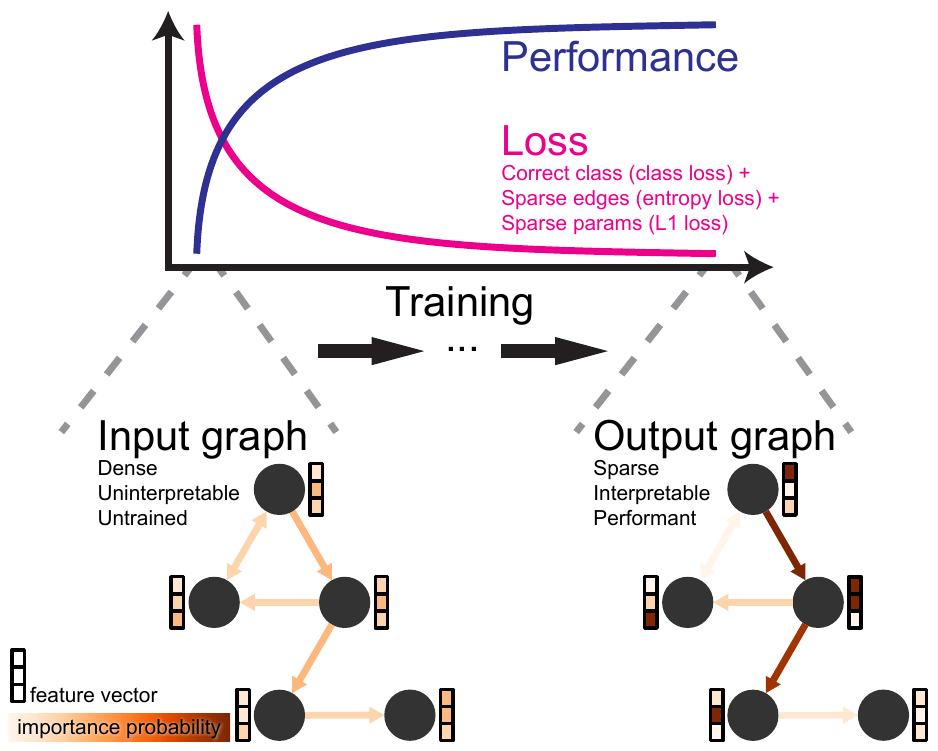}
    \label{fig:summaryfig}
\end{SCfigure}

Our work seeks to build upon the strengths of inherent, graph-specific, and post-hoc interpretability modeling methods. We present a model-agnostic framework with a component of inherent interpretability for defining important graph structure and node features, \underline{G}raph neural networks \underline{I}ncluding \underline{S}par\underline{S}e in\underline{T}erpretability (GISST; Fig \ref{fig:summaryfig}). GISST can be used for explaining predictions of any GNN model, over any graph (homogeneous or heterogeneous), for any graph modeling task (e.g. link prediction, node classification, or community detection), and for single and multiple instances (i.e. for a single entity or across all entities of the same class, respectively). This is achieved through a single self-attention layer that learns an importance probability for each node feature and edge in the input graph. These probabilities are optimized during model training and thus tied to the task-specific performance. To further facilitate generation of an interpretable and explanatory subgraph from a larger, potentially noisy graph and featureset, GISST sparsifies input graph topology and features via entropy and L1 regularization of the importance probabilities. The resulting sparse probabilities can be further probed with and may better enable post-hoc interpretation methods.

As compared to other interpretation methods across an array of synthetic and real-world datasets, GISST achieves higher average edge explanation precision and better identifies known important graph structures, respectively. Specifically, on synthetic datasets, GISST recognizes known network motifs and node features responsible for node labels, outperforming alternative approaches by an average of up to 0.241 in explanation precision. On real-world data such as molecular graphs, GISST correctly identifies known toxic chemical atomic components and molecular motifs qualitatively better than other methods. In sum, our work provides accurate, appropriate subgraph interpretations for GNN models across an array of modeling tasks.

\section{Related work}

The problems related to interpretability of neural network-based or other "black box" algorithms have received great attention in the machine learning field \citep{rudin2019stop}. Proposed solutions span a continuum from performing post-hoc analyses on trained models to building inherently interpretable models.

Post-hoc analyses allow interpretation only after model training, as they rely on either generating simplified, interpretable approximations of the prediction decision boundary \citep{lakkaraju2017interpretable,Ribeiro_2016,Augasta2011ReverseET,Zilke2016DeepREDR}, or investigate the nature of the decision boundary \citep{simonyan2013deep,Erhan2009VisualizingHF, baehrens2009explain}. By their post-hoc nature, these approaches give windows into the complex function a given model has learned; since they are not part of the model learning process itself, they can be noisy, inaccurate, misleading, or otherwise fail to provide a satisfactorily comprehensive explanation or reason for how the model actually generated an output \citep{adebayo2018sanity, rudin2019stop}.

Efforts have been made to build inherently interpretable models that still possess the rich, complex representational power of black box neural network models. Some approaches seek to reduce the complexity of the model through model compression or parameter sparsification \citep{zhang2015ell1regularized,han2015deep,He_2018,hinton2015distilling}. Other approaches use attention-based models, whereby the learned attention weights that help generate model predictions are directly used to identify features that were important for the predictions \citep{velivckovic2017graph}. To further improve interpretability, other inherent methods seek to build and leverage a library of prototypical featuresets, or generate small decision trees representative of the decision boundary \citep{wu2017sparsity,chen2018looks}. Importantly,  these approaches provide interpretability as part of the model training process, such that the explanations provided were those that the model actually used in its computation.

Although graphs as data structures are inherently interpretable, computation graphs used in GNNs are often not; interpretability methods for graph-based models have been hitherto poorly explored. Many of the aforementioned approaches may not be well-suited or easily adapted for graphs. For instance, attention-based GNNs such as Graph Attention Networks (GAT) \citep{velivckovic2017graph} provide edge importance that lack specificity to a given node, precluding node-specific prediction explanations. Furthermore, each layer of GAT has a different set of attention weights, and it is not apparent how to aggregate these weights for a simple interpretation. One recent study addressed shortcomings of graph interpretations by developing a novel post-hoc interpretation method, GNNExplainer, usable on any graph \citep{ying2019gnnexplainer}. This method formulates the interpretation problem as an optimization task to maximize mutual information between a GNN prediction and a distribution of possible (inherently interpretable) subgraph structures. 

GISST incorporates a component of inherent interpretability as part of the model optimization process by generating importance probabilities for all edges and features. Unlike GNNExplainer, GISST therefore provides explanations that require no additional post-hoc training or hyperparameter optimization. In addition, GISST better facilitates interpretation and avoids computational instabilities associated with gradient-based post-hoc interpretations on graphs by yielding explanations taken with respect to edge or feature importance probabilities, rather than the adjacency matrix or feature itself, respectively.

\section{GISST}

\subsection{Problem formulation}

Let $G$ be a graph with $N$ nodes attributed with $d$-dimensional node features, $A \in \{0, 1\}^{N \times N}$ the adjacency matrix of $G$, and $X \in \mathbb{R}^{N \times d}$ the node feature matrix. For any given GNN model $\Phi$, the input $G$ and $X$ may not be entirely useful for the GNN task, especially when $G$ and $X$ are noisy. Therefore, we postulate that an important subgraph $G_s$ with adjacency matrix $A_s$, and an important subset of the node features $X_s$ contribute to the GNN performance. We can formulate a GNN task as a two-step process, where $G_s$ and $X_s$ are first generated and then used for computing the output of $\Phi$. The optimal $G_s$ and $X_s$ therefore can provide importance interpretation for $\Phi$.

The inference of $G_s$ by exploring all possible subgraphs of $G$ is not computationally feasible as the problem is NP-hard \citep{chandrasekaran2008complex}, and an exhaustive search for $X_s$ is computationally expensive for GNNs. Therefore, we consider $G_s$ and $X_s$ as random variables. Specifically, we take the approach of mean field approximation and model
\begin{equation} \label{eq:as_dist}
    \mathbb{I}\{A_s[i,j]\ \ne 0 \} \sim_{ind} Bernoulli(p_{i,j})
\end{equation}
with the graph topology constraint that $p_{i,j} \le A[i,j]$ for all $i,j$. Here, $\mathbb{I}\{A_s[i,j]\ \ne 0\}$ denotes an indicator random variable for the inclusion of edge $(i, j)$ in the important subgraph $G_s$, and $p_{i,j}$ is the probability of the inclusion. Similarly, we model the importance of each node feature through
\begin{equation} \label{eq:xs_dist}
    \mathbb{I}\big\{X_s[k,l] \ne 0\big\} \sim_{ind} Bernoulli(p_{k,l})
\end{equation}
for all $k,l$, where the $l$th feature has a probability of $p_{k,l}$ to be important for node $k$. Let $P_{A_s} \in [0, 1]^{N \times N}$ be the probability matrix for $A_s$, where $P_{A_s}[i,j] = p_{i,j}$ for all $i,j$. Similarly, let $P_{X_s}$ denote the probability matrix for $X_s$. Then the expected output of a given GNN model $\Phi$ can be expressed as
\begin{equation} \label{eq:exp_out_1}
    o = \Phi(A \odot P_{A_s}, X \odot P_{X_s})
\end{equation}
where $\odot$ denotes the Hadamard element-wise product. Here, we translate the problem of identifying $G_s$ and $X_s$ into inference of the importance probability $P_{A_s}$ and $P_{X_s}$. This inference of a dynamic graph through a probabilistic approach is also found in previous work \citep{molinelli2013perturbation, kipf2018neural, ying2019gnnexplainer}.

\subsection{GISST model parameterization}
The probability matrix $P_{A_s}$ and $P_{X_s}$ provide the interpretation for a GNN model $\Phi$. To generalize the interpretation for unseen nodes, edges, and node features, we parameterize $P_{A_s}$ and $P_{X_s}$ in GISST. We assume that each node feature has the same importance probability across all the nodes, that is $P_{X_s}[k,l] = p_l$ for all $k$. For optimization, each probability $p_l$ is parameterized by $m_l \in \mathbb{R}$ through $\sigma(m_l)$, where $\sigma$ denotes the sigmoid function.

To compute $P_{A_s}$, we apply a single-layer feed-forward network to the source and target node features, similar to the attention mechanism in GAT \citep{velivckovic2017graph}. Specifically, we have
\begin{equation} \label{eq:att_mech}
    P_{A_s}[i,j] = \sigma\Big([x_i \odot p || x_j \odot p]^T b\Big)
\end{equation}
where $p$ is a vector containing the node feature probabilities, $b$ is a parameter to be learned, and $||$ denotes concatenation. Note that the node feature values are adjusted for their feature importance in this computation. This is to remove noisy, unimportant node features when computing the edge importance probability. Overall, the parameters $m \in \mathbb{R}^d$ containing $m_l$ and $b \in \mathbb{R}^{2d}$ are used for the inference of $P_{A_s}$ and $P_{X_s}$.

\subsection{GISST loss function}
To optimize the GISST parameters $m$ and $b$, we consider the output computation in Equation \ref{eq:exp_out_1}, which is used in the model prediction loss function. GISST can be used for any graph-related task such as node classification, link prediction, and community detection. Without loss of generality, here we focus on the task of node classification as a concrete demonstration.

Suppose we have possible labels $\mathcal{Y} \in \{0, 1, ..., C\}$. Let $\hat{y} \in (0,1)^{N \times C}$ be the softmax-transformed output of $\Phi(A \odot P_{A_s}, X \odot P_{X_s})$. Then we can consider the cross-entropy loss function
\begin{equation} \label{eq:loss_class}
        \mathcal{L}_{class} = \frac{1}{N} \sum_{i=1}^{N} \sum_{c=1}^{C} -y_{i,c} \log{(\hat{y}_{i,c})}
\end{equation}
where $y$ is the one-hot encoding for the actual node labels. Because the GISST parameters $m$ and $b$ are included in the classification loss through $P_{A_s}$ and $P_{X_s}$, GISST has the desirable property that it infers $P_{A_s}$ and $P_{X_s}$ specifically important for a GNN model's classification performance.

To induce sparsity on $P_{A_s}$, we consider the L1 and entropy regularization loss
\begin{equation} \label{eq:loss_edge_l1}
    \mathcal{L}_{P_{A_s}, L1} = \frac{1}{N^2} \sum_{i,j} P_{A_s}[i,j] 
\end{equation}
and
\begin{equation} \label{eq:loss_edge_ent}
    \mathcal{L}_{P_{A_s}, entropy} = \frac{1}{N^2} \sum_{i,j} P_{A_s}[i,j]\log(P_{A_s}[i,j]) + (1 - P_{A_s}[i,j])\log(1 - P_{A_s}[i,j])
\end{equation}
Both the L1 and entropy regularization terms induce sparsity by driving some entries in $P_{A_s}$ towards zero. At the same time, the entropy loss also induces important edges to have probability close to one. The same L1 and entropy regularization are applied to $P_{X_s}$, denoted as $\mathcal{L}_{P_{X_s}, L1}$ and $\mathcal{L}_{P_{X_s}, entropy}$

Finally, to optimize a GISST model, the overall loss function is the sum of $\mathcal{L}_{class}$, $\mathcal{L}_{P_{A_s}, L1}$, $\mathcal{L}_{P_{A_s}, entropy}$, $\mathcal{L}_{P_{X_s}, L1}$, and $\mathcal{L}_{P_{X_s}, entropy}$, with penalty coefficients for the regularization losses as hyperparameters. Note that these hyperparameters can be tuned for classification performance. Hence the induced sparsity is useful for the particular classification task.

\subsection{GISST interpretability}
After optimizing a GISST model, the importance probabilities in $P_{A_s}$ and $P_{X_s}$ are used for interpretation. An important edge would correspond to a high probability value in $P_{A_s}$, similarly for an important node feature with $P_{X_s}$. For identifying important subgraph and node features of a specific classification instance, we can interpret the gradient of the instance-specific predicted loss with respect to $P_{A_s}$ and $P_{X_s}$. For a group of instances, their average predicted loss can be used for interpretation.

\subsection{GISST extensions}

\paragraph{Any GNN model.} The importance probability $P_{A_s}$ and $P_{X_s}$ can be seen as weighted adjacency matrix and node feature weights. Therefore, GISST can be applied to any GNN model that accepts weighted adjacency matrix and node feature matrix as the input. This includes Graph Convolutional Networks (GCN) \citep{kipf2016semi}, GNN with convolution filters \citep{bianchi2019graph, defferrard2016convolutional}, and Gated Graph Sequence Neural Networks \citep{li2015gated}, to name a few. Furthermore, the classification loss in the overall GISST loss can be substituted for any graph-related loss function. Therefore, we can conceptualize GISST as a general framework that is model-agnostic and task-agnostic.

\paragraph{Integrating edge attribute importance.} Here we demonstrate that, in theory, the importance of edge attributes such as edge features and edge types can be assessed in GISST. The key insight is that the attention mechanism in Equation \ref{eq:att_mech} is equivalent to fitting a logistic regression model for the importance of an edge. Additional features can be included in the logistic regression model, and their importance can be interpreted through the learned coefficients. For example, suppose we have edge features $z_{i,j} \in \mathbb{R}^{h}$ for all $i,j$, then we can reformulate Equation \ref{eq:att_mech} to obtain
\begin{equation} \label{eq:edge_attrs}
    P_{A_s}[i,j] = \sigma\Big([x_i \odot p || x_j \odot p]^T b + z_{i,j}^T a\Big)
\end{equation}
where $a \in \mathbb{R}^{h}$ contains the edge feature coefficients, which are used for interpretation as in a logistic regression model. Edge types can be incorporated as discrete edge features. Alternatively, a distinct set of the parameters $b$ and $a$ can be used for each edge type. That is
\begin{equation} \label{eq:edge_type}
    P_{A_s}[i,j,r] = \sigma\Big([x_i \odot p || x_j \odot p]^T b_r + z_{i,j}^T a_r\Big)
\end{equation}
where $r$ indexes the edge types. With these modifications, GISST can be applied to edge-attributed GNN models such as Relational Graph Convolutional Networks \citep{schlichtkrull2018modeling}.

\section{Experiments}

In this section, we describe the graph datasets used in our experiments, the quantitative metric used for evaluating important subgraph and node feature identification, alternative methods for comparison, and experimental details. The quantitative results on synthetic datasets and qualitative results on real-world datasets demonstrate that GISST can identify important, meaningful subgraph and node features.

\subsection{Datasets}

\paragraph{Synthetic datasets.} Four synthetic datasets for node classification are constructed with ground truth edge and node feature importance to quantitatively evaluate important edge and node feature identification performance. The graph structures are generated in accordance to \citep{ying2019gnnexplainer}. However, to evaluate important node feature identification, we generate 40 important and 10 unimportant node features from Gaussian distributions. In order to differentiate between our synthetic datasets and the ones in \citep{ying2019gnnexplainer}, we describe our synthetic datasets as "noisy." Overall, we have the following. (1) Noisy BA-House consists of a Barabasi-Albert (BA) graph of 300 nodes with 80 five-node house-structured motifs, where the nodes in the BA graph belongs to one class, and the top, middle, and bottom of a house motif belong to three distinct classes. (2) Noisy BA-Community is a union of two Noisy BA-House graphs joined by random edges, with 8 classes based on graph membership and structural roles. (3) Noisy Tree-Cycle consists of an 8-level balanced binary tree with 80 six-node cycle motifs, where nodes in the tree and cycles are considered as two different classes. (4) Noisy Tree-Grid is the same as Noisy Tree-Cycle except that the motif structure is a 3-by-3 grid. The edges defining a motif structure such as a house or cycle are considered important for the nodes in the structure. Important node features are generated with a mean difference of 1 between different classes, whereas unimportant node features are generated from the same distribution. A standard deviation of 0.15 is used for datasets containing more than two classes, and 0.5 for binary classification datasets.

\paragraph{Real-world datasets.} To visually assess importance interpretation in real-world datasets, we consider two graph classification datasets. (1) The Mutagenicity dataset contains 4,337 molecule graphs labeled as mutagenic or non-mutagenic in \textit{S. typhimurium} \citep{debnath1991structure}. (2) The REDDIT-BINARY dataset contains 2,000 graphs labeled as question/answer-based or discussion-based community in the content-aggregation website Reddit \citep{yanardag2015deep}.

\subsection{Evaluation}

In synthetic datasets where we know the ground truth edge importance, we can quantitatively evaluate important edge identification performance. In real-world applications, only a few most important nodes and edges are visualized for human-readable interpretation and cost-associated actionable impact. We simulate this scenario and focus on edge explanation precision, which is defined as the proportion of important edges in an important subgraph extracted based on edge importance scores. Important structures in the considered datasets are connected subgraphs with some minimum number of nodes $V_M$. Therefore, an important subgraph is extracted by a search to find the maximum threshold for removing less important edges, such that the subgraph contains at least $V_M$ nodes. This subgraph extraction procedure is the same as in \citep{ying2019gnnexplainer}. Node feature explanation precision is defined as the proportion of ground truth important node features in the top-K identified node features.

\subsection{Alternative methods}

We consider three alternative methods for generating edge and node feature importance score. (1) Average attention weight across the layers in a GAT model is computed as edge importance \citep{velivckovic2017graph}. GAT does not compute attention weights for node features, so the attention-based method is not applicable to node feature importance. (2) Gradients of the predicted classification loss with respect to the adjacency matrix and input node features are computed as edge importance and node feature importance, respectively. (3) GNNExplainer is a post-hoc model that requires further optimization to generate edge and node feature importance for a specific node \citep{ying2019gnnexplainer}.

\subsection{Implementation details}

All datasets are split 80/10/10 as the training, validation, and test set. Normalized node degree is used as node feature in the REDDIT-BINARY dataset. All the parameters are initialized using the Xavier uniform initialization \citep{glorot2010understanding}, and the Adam optimizer is used for training \citep{kingma2014adam}. For synthetic datasets, we set $V_M$ to be the size of the motifs, and the top-40 important node features are considered for evaluation. For the real-world datasets, $V_M$ is set to 15 to capture important complex structure such as fused carbon rings \citep{Kazius_2005}. Important subgraphs for the motif nodes and important node features for all the nodes in the test sets are extracted for evaluation. Only the computation subgraph, as opposed to the entire input graph, for a particular node is considered for subgraph extraction, since all the other information is not used for computing the node's output. For example, in a GNN with three graph convolution layers, only the three-step neighborhood of a node is considered. Average node feature importance score across nodes is used for node feature explanation precision, where the minmax normalization is applied to account for scale variability. For graph-level explanation, the average predicted loss across all the nodes is used for the gradient-based results and GISST. For GNNExplainer, the median of the edge importance, which is robust to outliers, is used. To simulate realistic settings where the ground truth edge and node feature importance labels are not known prior to explanation evaluation, the GNNExplainer hyperparameters are not tuned and are selected based on the suggestions in \citep{ying2019gnnexplainer}. Code, datasets, and further optimization details can be found in the Appendix. All the models are implemented in PyTorch and PyTorch Geometric \citep{paszke2019pytorch, fey2019fast}.

\subsection{Results}

\paragraph{Results for synthetic datasets.} Edge explanation results on the synthetic datasets are shown in Table \ref{tab:syn_edge_precision}. Overall, GISST has an average edge explanation precision of 0.772 across all the synthetic datasets, 0.047 higher than the gradient-based method (Grad), 0.204 higher than GNNExplainer, and 0.241 higher than GAT. GISST also has the highest average edge explanation accuracy and similar recall performance as compared to the other methods (see Appendix).

Node feature explanation results are shown in Table \ref{tab:syn_node_feat_precision}. For the tree-based binary classification datasets, GISST performs the best in terms of node feature explanation precision. Overall, GISST has the highest average performance of 0.825 across the synthetic datasets. GISST also has the highest average node feature explanation accuracy and recall performance (see Appendix).

\begin{table*}[h]
  \caption{Edge explanation precision results in synthetic datasets.}
  \label{tab:syn_edge_precision}
  \centering
  \addtolength{\tabcolsep}{-4pt}
  \begin{tabular}{lcccc}
    \toprule
    \textbf{Method}  & \textbf{Noisy BA-House}  & \textbf{ Noisy BA-Comm}  & \textbf{Noisy Tree-Cycle}  & \textbf{Noisy Tree-Grid} \\
    \midrule
    GAT              & 0.588              & 0.213                 & 0.715                & 0.609             \\
    Grad             & 0.650              & 0.539                 & 0.858                & 0.853             \\
    GNNExplainer     & 0.324              & \textbf{0.583}        & 0.613                & 0.753             \\
    GISST            & \textbf{0.708}     & 0.576                 & \textbf{0.917}       & \textbf{0.886}    \\
    \bottomrule
  \end{tabular}
\end{table*}

\begin{table*}[h]
  \caption{Node feature explanation precision results in synthetic datasets.}
  \label{tab:syn_node_feat_precision}
  \centering
  \addtolength{\tabcolsep}{-4pt}
  \begin{tabular}{lcccc}
    \toprule
    \textbf{Method}  & \textbf{Noisy BA-House}  & \textbf{ Noisy BA-Comm}  & \textbf{Noisy Tree-Cycle}  & \textbf{Noisy Tree-Grid} \\
    \midrule
    GAT              & NA                      & NA                        & NA                   & NA                 \\
    Grad             & \textbf{0.800}          & \textbf{0.800}            & 0.750                & 0.850              \\
    GNNExplainer     & 0.750                   & 0.775                     & 0.825                & 0.800              \\
    GISST            & 0.750                   & 0.750                     & \textbf{0.875}       & \textbf{0.925}     \\
    \bottomrule
  \end{tabular}
\end{table*}

\paragraph{Results for real-world datasets.} We also qualitatively assess model interpretation in real-world datasets. We first note that the graph classification performance is similar across GAT, GCN, and GISST (Table \ref{tab:real_graph_acc}).

\begin{table*}[h]
  \caption{Graph classification accuracy results for real-world datasets.}
  \label{tab:real_graph_acc}
  \centering
  \addtolength{\tabcolsep}{+6pt}
  \begin{tabular}{lcc}
    \toprule
    \textbf{Method}  & \textbf{Mutagenicity}  & \textbf{REDDIT-BINARY} \\
    \midrule
    GCN              & 0.777                 & \textbf{0.920}        \\
    GAT              & \textbf{0.804}        & \textbf{0.920}        \\
    GISST            & 0.786                 & 0.915                 \\
    \bottomrule
  \end{tabular}
\end{table*}

\begin{figure}[h]
    \centering
    \includegraphics[scale=0.48]{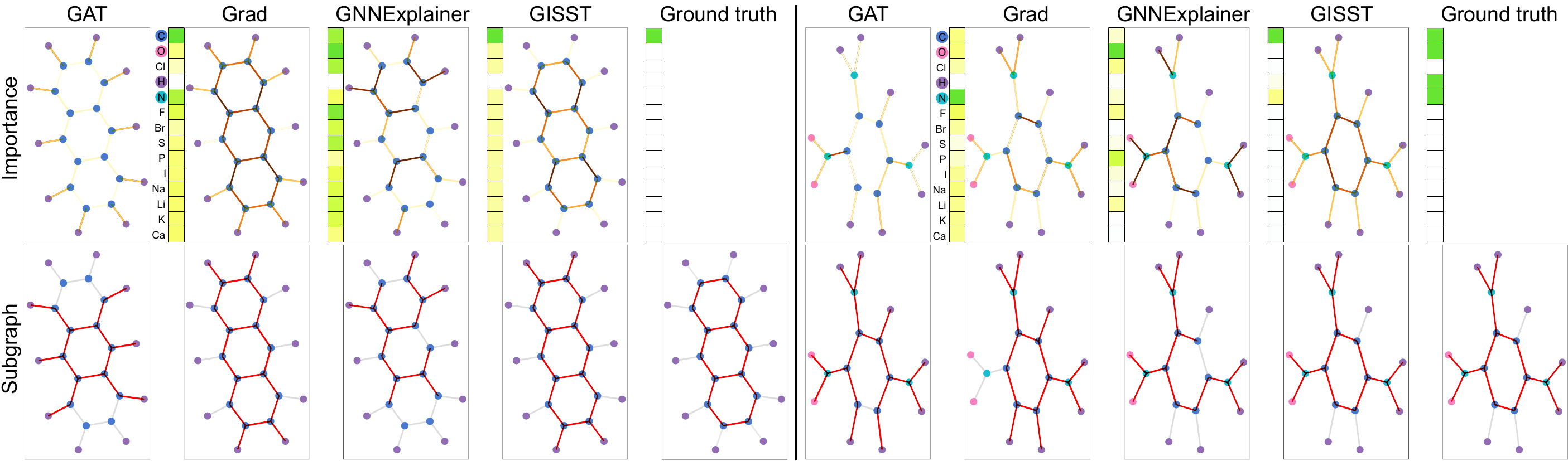}
    \caption{Edge or node feature importance scores ("Importance") and prediction explanation subgraphs ("Subgraph") for two representative mutagenic molecule graphs from Mutagenicity dataset. Edge "Importance" corresponds to attention weights (GAT), gradients of the predicted loss function with respect to the adjacency matrix, $A$ (Grad), explanation weights (GNNExplainer), or gradients of the predicted loss function with respect to the edge probability, $P_{A_s}$ (GISST). Node feature importance not applicable for GAT. Ground truth subgraph is defined as in \citep{Kazius_2005}.}
    \label{fig:mutag}
\end{figure}

In the Mutageniticity dataset, we observe that GISST and Grad can correctly identify fused carbon rings as mutagenic structure \citep{Kazius_2005}, whereas GNNExplainer and GAT tend to miss some carbon-carbon bonds (Fig \ref{fig:mutag}). Additionally, GISST can identify the known mutagenic aromatic nitro and amine groups (i.e. the complete carbon ring, $NO_2$, and $NH_2$ groups), whereas other methods tend to have an incomplete ring or missing a functional group. GISST also selectively identifies the most important atoms, whereas other methods may place higher importance on atoms absent in the molecules (e.g. $F$ and $Br$).

\begin{figure}[h]
    \centering
    \includegraphics[scale=0.5]{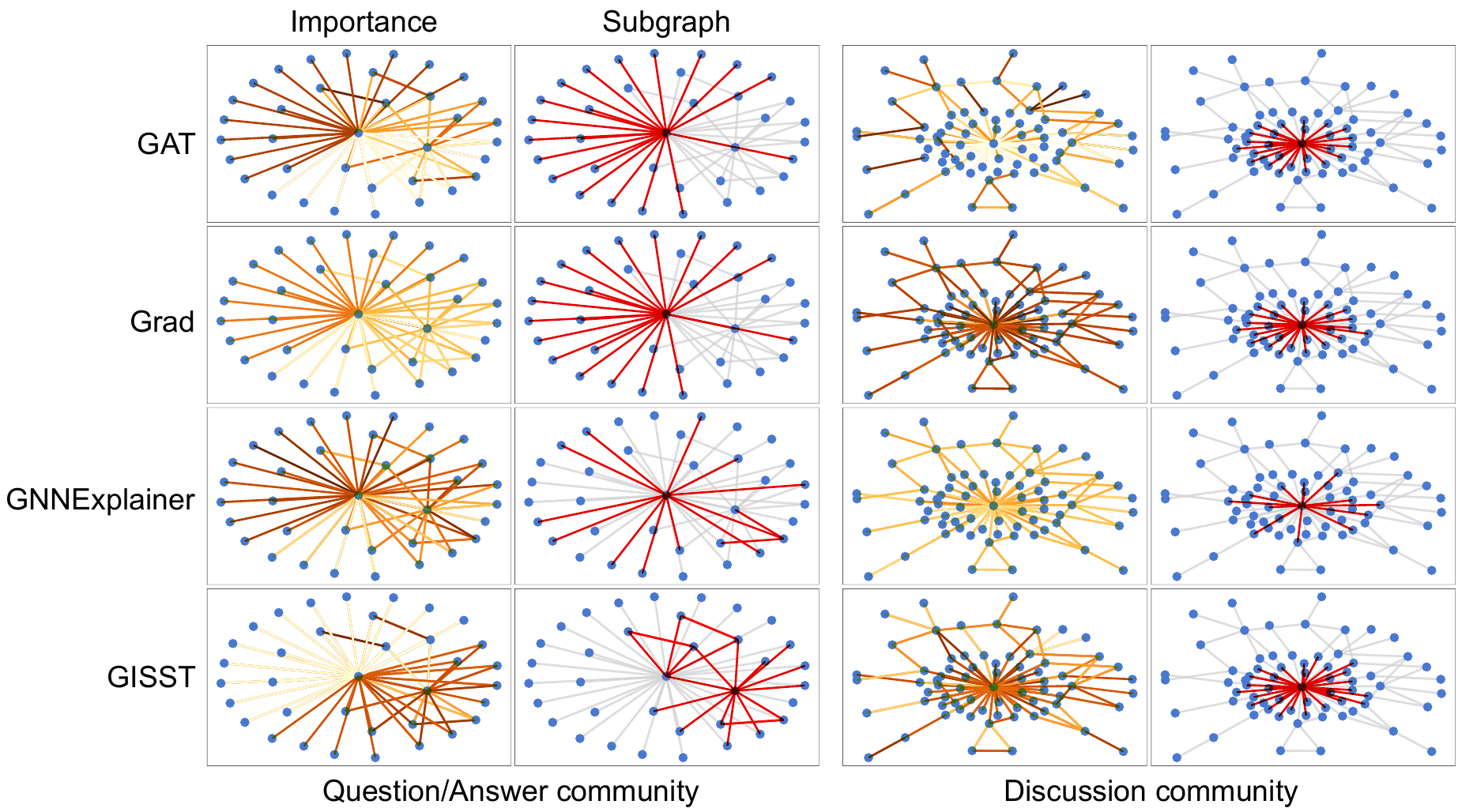}    
    \caption{Edge importance scores ("Importance") and prediction explanation subgraphs ("Subgraph") for two representative REDDIT-BINARY community graphs. "Importance" is computed for each model as in Fig \ref{fig:mutag}.}
    \label{fig:reddit}
\end{figure}

In the REDDIT-BINARY dataset, GISST and GNNExplainer can identify the two experts with high degrees (i.e. answering multiple questions) in a question/answer-based community, whereas Grad and GAT tend to focus on only one expert (Fig \ref{fig:reddit}).

\section{Conclusion}

We developed GISST, a graph model-agnostic and task-agnostic method that yields a sparse featureset and a subgraph structure as part of the model training (optimization) process. As such, GISST provides a component of inherent interpretability by facilitating identification of explanatory subgraphs, thus aiding human understanding of relations and rules inherent in the data and predictive of some outcome. Across synthetic and real-world datasets and both graph-level or node-level prediction problems, our model provides predictive performance comparable to traditional Graph Neural Network formulations, while additionally identifying important features and edges based on importance probabilities that quantitatively and qualitatively match or exceed existing post-hoc interpretation methods. 

Future work could explore alternate means of inferring feature ($P_{X_s}$) and edge ($P_{A_s}$) probabilities to improve interpretable graph generation, as well as explore end-to-end learning of the important subgraph sizes. Further, as application of graph interpretability methods is still nascent, we anticipate utility of our method for other graph neural network architectures and vastly different real-world datasets, which will both highlight the power of our approach as well as additional opportunities for improvement.

\medskip
\small

\section*{Broader Impact}

Our work seeks to address the ongoing issue of model interpretability and explainability\textemdash that is, broadly, knowing what human-interpretable information (e.g. in the input data or learned model parameters) lead to a particular model output \citep{rudin2019stop}. We present here one solution relevant to graph-based neural network models, which should have immediate impact for low-stakes decisions (e.g. recommendation engines, generating new scientific hypotheses), and should be used with mindfulness of any potential limitations if applied to high-stakes decisions (e.g. healthcare, policy). 

Finally, it is beyond the scope of our method to correct for any intrinsic data biases leading to interpretations or explanations that could reinforce harmful stereotypes or preconceived notions. That said, a positive benefit of our approach is that it may enable the identification of such biases by providing model output interpretations or explanations.

\section*{Acknowledgements}

We thank the entire Early Computational Oncology group at AstraZeneca, especially the Knowledge Graphs, Data Science, and Bioinformatics working groups, for their invaluable feedback and discussion.

\bibliography{arxiv_2020}

\begin{thebibliography}{10}

\bibitem{Ying_2018}
Rex Ying, Ruining He, Kaifeng Chen, Pong Eksombatchai, William~L. Hamilton, and
  Jure Leskovec.
\newblock Graph convolutional neural networks for web-scale recommender
  systems.
\newblock {\em Proceedings of the 24th ACM SIGKDD International Conference on
  Knowledge Discovery \& Data Mining}, Jul 2018.

\bibitem{Kovacs2019}
I.A. Kov{\'a}cs, K.~Luck, K.~Spirohn, and et~al.
\newblock Network-based prediction of protein interactions.
\newblock {\em Nat Commun}, 10:1240, 2019.

\bibitem{Zitnik_2018}
Marinka Zitnik, Monica Agrawal, and Jure Leskovec.
\newblock Modeling polypharmacy side effects with graph convolutional networks.
\newblock {\em Bioinformatics}, 34(13):i457–i466, Jun 2018.

\bibitem{Goh8685}
Kwang-Il Goh, Michael~E. Cusick, David Valle, Barton Childs, Marc Vidal, and
  Albert-L{\'a}szl{\'o} Barab{\'a}si.
\newblock The human disease network.
\newblock {\em Proceedings of the National Academy of Sciences},
  104(21):8685--8690, 2007.

\bibitem{gysi2020network}
Deisy~Morselli Gysi, Ítalo Do~Valle, Marinka Zitnik, Asher Ameli, Xiao Gan,
  Onur Varol, Helia Sanchez, Rebecca~Marlene Baron, Dina Ghiassian, Joseph
  Loscalzo, and Albert-László Barabási.
\newblock Network medicine framework for identifying drug repurposing
  opportunities for covid-19, 2020.

\bibitem{hamilton2017representation}
William~L. Hamilton, Rex Ying, and Jure Leskovec.
\newblock Representation learning on graphs: Methods and applications, 2017.

\bibitem{you2018graph}
Jiaxuan You, Bowen Liu, Rex Ying, Vijay Pande, and Jure Leskovec.
\newblock Graph convolutional policy network for goal-directed molecular graph
  generation, 2018.

\bibitem{xu2018powerful}
Keyulu Xu, Weihua Hu, Jure Leskovec, and Stefanie Jegelka.
\newblock How powerful are graph neural networks?, 2018.

\bibitem{battaglia2018relational}
Peter~W. Battaglia, Jessica~B. Hamrick, Victor Bapst, Alvaro Sanchez-Gonzalez,
  Vinicius Zambaldi, Mateusz Malinowski, Andrea Tacchetti, David Raposo, Adam
  Santoro, Ryan Faulkner, Caglar Gulcehre, Francis Song, Andrew Ballard, Justin
  Gilmer, George Dahl, Ashish Vaswani, Kelsey Allen, Charles Nash, Victoria
  Langston, Chris Dyer, Nicolas Heess, Daan Wierstra, Pushmeet Kohli, Matt
  Botvinick, Oriol Vinyals, Yujia Li, and Razvan Pascanu.
\newblock Relational inductive biases, deep learning, and graph networks, 2018.

\bibitem{Goyal_2018}
Palash Goyal and Emilio Ferrara.
\newblock Graph embedding techniques, applications, and performance: A survey.
\newblock {\em Knowledge-Based Systems}, 151:78–94, Jul 2018.

\bibitem{rudin2019stop}
Cynthia Rudin.
\newblock Stop explaining black box machine learning models for high stakes
  decisions and use interpretable models instead.
\newblock {\em Nature Machine Intelligence}, 1(5):206--215, 2019.

\bibitem{lakkaraju2017interpretable}
Himabindu Lakkaraju, Ece Kamar, Rich Caruana, and Jure Leskovec.
\newblock Interpretable \& explorable approximations of black box models, 2017.

\bibitem{Ribeiro_2016}
Marco~Tulio Ribeiro, Sameer Singh, and Carlos Guestrin.
\newblock “why should i trust you?”.
\newblock {\em Proceedings of the 22nd ACM SIGKDD International Conference on
  Knowledge Discovery and Data Mining}, Aug 2016.

\bibitem{Augasta2011ReverseET}
M.~Gethsiyal Augasta and T.~Kathirvalavakumar.
\newblock Reverse engineering the neural networks for rule extraction in
  classification problems.
\newblock {\em Neural Processing Letters}, 35:131--150, 2011.

\bibitem{Zilke2016DeepREDR}
Jan~Ruben Zilke, Eneldo~Loza Menc{\'i}a, and Frederik Janssen.
\newblock Deepred - rule extraction from deep neural networks.
\newblock In {\em DS}, 2016.

\bibitem{simonyan2013deep}
Karen Simonyan, Andrea Vedaldi, and Andrew Zisserman.
\newblock Deep inside convolutional networks: Visualising image classification
  models and saliency maps, 2013.

\bibitem{Erhan2009VisualizingHF}
Dumitru Erhan, Yoshua Bengio, Aaron~C. Courville, and Pascal Vincent.
\newblock Visualizing higher-layer features of a deep network.
\newblock 2009.

\bibitem{baehrens2009explain}
David Baehrens, Timon Schroeter, Stefan Harmeling, Motoaki Kawanabe, Katja
  Hansen, and Klaus-Robert Mueller.
\newblock How to explain individual classification decisions, 2009.

\bibitem{adebayo2018sanity}
Julius Adebayo, Justin Gilmer, Michael Muelly, Ian Goodfellow, Moritz Hardt,
  and Been Kim.
\newblock Sanity checks for saliency maps, 2018.

\bibitem{zhang2015ell1regularized}
Yuchen Zhang, Jason~D. Lee, and Michael~I. Jordan.
\newblock $\ell_1$-regularized neural networks are improperly learnable in
  polynomial time, 2015.

\bibitem{han2015deep}
Song Han, Huizi Mao, and William~J. Dally.
\newblock Deep compression: Compressing deep neural networks with pruning,
  trained quantization and huffman coding, 2015.

\bibitem{He_2018}
Yihui He, Ji~Lin, Zhijian Liu, Hanrui Wang, Li-Jia Li, and Song Han.
\newblock Amc: Automl for model compression and acceleration on mobile devices.
\newblock {\em Lecture Notes in Computer Science}, page 815–832, 2018.

\bibitem{hinton2015distilling}
Geoffrey Hinton, Oriol Vinyals, and Jeff Dean.
\newblock Distilling the knowledge in a neural network, 2015.

\bibitem{velivckovic2017graph}
Petar Veli{\v{c}}kovi{\'c}, Guillem Cucurull, Arantxa Casanova, Adriana Romero,
  Pietro Lio, and Yoshua Bengio.
\newblock Graph attention networks.
\newblock {\em arXiv preprint arXiv:1710.10903}, 2017.

\bibitem{wu2017sparsity}
Mike Wu, Michael~C. Hughes, Sonali Parbhoo, Maurizio Zazzi, Volker Roth, and
  Finale Doshi-Velez.
\newblock Beyond sparsity: Tree regularization of deep models for
  interpretability, 2017.

\bibitem{chen2018looks}
Chaofan Chen, Oscar Li, Chaofan Tao, Alina~Jade Barnett, Jonathan Su, and
  Cynthia Rudin.
\newblock This looks like that: Deep learning for interpretable image
  recognition, 2018.

\bibitem{ying2019gnnexplainer}
Zhitao Ying, Dylan Bourgeois, Jiaxuan You, Marinka Zitnik, and Jure Leskovec.
\newblock Gnnexplainer: Generating explanations for graph neural networks.
\newblock In {\em Advances in Neural Information Processing Systems 32}, pages
  9244--9255. Curran Associates, Inc., 2019.

\bibitem{chandrasekaran2008complex}
Venkat Chandrasekaran, Nathan Srebro, and Prahladh Harsha.
\newblock Complexity of inference in graphical models.
\newblock In {\em Proceedings of the Twenty-Fourth Conference on Uncertainty in
  Artificial Intelligence}, UAI’08, page 70–78, Arlington, Virginia, USA,
  2008. AUAI Press.

\bibitem{molinelli2013perturbation}
Evan~J Molinelli, Anil Korkut, Weiqing Wang, Martin~L Miller, Nicholas~P
  Gauthier, Xiaohong Jing, Poorvi Kaushik, Qin He, Gordon Mills, David~B Solit,
  et~al.
\newblock Perturbation biology: inferring signaling networks in cellular
  systems.
\newblock {\em PLoS computational biology}, 9(12), 2013.

\bibitem{kipf2018neural}
Thomas Kipf, Ethan Fetaya, Kuan-Chieh Wang, Max Welling, and Richard Zemel.
\newblock Neural relational inference for interacting systems.
\newblock {\em arXiv preprint arXiv:1802.04687}, 2018.

\bibitem{kipf2016semi}
Thomas~N Kipf and Max Welling.
\newblock Semi-supervised classification with graph convolutional networks.
\newblock {\em arXiv preprint arXiv:1609.02907}, 2016.

\bibitem{bianchi2019graph}
Filippo~Maria Bianchi, Daniele Grattarola, Cesare Alippi, and Lorenzo Livi.
\newblock Graph neural networks with convolutional arma filters.
\newblock {\em arXiv preprint arXiv:1901.01343}, 2019.

\bibitem{defferrard2016convolutional}
Micha{\"e}l Defferrard, Xavier Bresson, and Pierre Vandergheynst.
\newblock Convolutional neural networks on graphs with fast localized spectral
  filtering.
\newblock In {\em Advances in neural information processing systems}, pages
  3844--3852, 2016.

\bibitem{li2015gated}
Yujia Li, Daniel Tarlow, Marc Brockschmidt, and Richard Zemel.
\newblock Gated graph sequence neural networks.
\newblock {\em arXiv preprint arXiv:1511.05493}, 2015.

\bibitem{schlichtkrull2018modeling}
Michael Schlichtkrull, Thomas~N Kipf, Peter Bloem, Rianne Van Den~Berg, Ivan
  Titov, and Max Welling.
\newblock Modeling relational data with graph convolutional networks.
\newblock In {\em European Semantic Web Conference}, pages 593--607. Springer,
  2018.

\bibitem{debnath1991structure}
Asim~Kumar Debnath, Rosa~L Lopez~de Compadre, Gargi Debnath, Alan~J Shusterman,
  and Corwin Hansch.
\newblock Structure-activity relationship of mutagenic aromatic and
  heteroaromatic nitro compounds. correlation with molecular orbital energies
  and hydrophobicity.
\newblock {\em Journal of medicinal chemistry}, 34(2):786--797, 1991.

\bibitem{yanardag2015deep}
Pinar Yanardag and SVN Vishwanathan.
\newblock Deep graph kernels.
\newblock In {\em Proceedings of the 21th ACM SIGKDD International Conference
  on Knowledge Discovery and Data Mining}, pages 1365--1374, 2015.

\bibitem{glorot2010understanding}
Xavier Glorot and Yoshua Bengio.
\newblock Understanding the difficulty of training deep feedforward neural
  networks.
\newblock In {\em Proceedings of the thirteenth international conference on
  artificial intelligence and statistics}, pages 249--256, 2010.

\bibitem{kingma2014adam}
Diederik~P Kingma and Jimmy Ba.
\newblock Adam: A method for stochastic optimization.
\newblock {\em arXiv preprint arXiv:1412.6980}, 2014.

\bibitem{Kazius_2005}
Jeroen Kazius, Ross McGuire, and Roberta Bursi.
\newblock Derivation and validation of toxicophores for mutagenicity
  prediction.
\newblock {\em Journal of Medicinal Chemistry}, 48(1):312--320, 2005.
\newblock PMID: 15634026.

\bibitem{paszke2019pytorch}
Adam Paszke, Sam Gross, Francisco Massa, Adam Lerer, James Bradbury, Gregory
  Chanan, Trevor Killeen, Zeming Lin, Natalia Gimelshein, Luca Antiga, et~al.
\newblock Pytorch: An imperative style, high-performance deep learning library.
\newblock In {\em Advances in Neural Information Processing Systems}, pages
  8024--8035, 2019.

\bibitem{fey2019fast}
Matthias Fey and Jan~Eric Lenssen.
\newblock Fast graph representation learning with pytorch geometric.
\newblock {\em arXiv preprint arXiv:1903.02428}, 2019.

\end{thebibliography}

\beginsupplement

\section{Further implementation details}
For each dataset, the following models are trained: a GAT model, a GCN model (for the gradient-based method and GNNExplainer), and a GISST model (with GCN as the classifier). Two fully connected layers are used with mean pooling to generate graph-level prediction for graph classification datasets. For each model, we tune the number of hidden units $\in \{16, 32\}$, number of hidden graph convolution/attention layer $\in \{2, 3, 4\}$, and learning rate $\in \{0.001, 0.005\}$. Overall L2 penalty $\in \{0.0005, 0.005, 0.005, 0.5\}$ is tuned for GCN and GAT, and $\{0.0005, 0.005\}$ for GISST. Dropout rate across all hidden layers $\in \{0.1, 0.2, 0.3, 0.4\}$ is tuned for GCN and GAT, and $\{0.1, 0.2\}$ for GISST. The L1 penalty $\in \{0.005, 0.05, 0.5\}$ for GISST edge probability and $\{0.0005, 0.005, 0.05\}$ for node feature probability are tuned. The entropy penalty $\in \{0.01, 0.1, 1.0\}$  for GISST edge probability and $\{0.001, 0.01, 0.1\}$ for node feature probability are tuned. Models are trained for 1000 epochs for the synthetic datasets and 500 epochs for the real-world datasets. For the BA-based synthetic datasets, the final models have 3 graph convolution/attention layers, 4 graph layers for the tree-based synthetic datasets, and 3 graph layers for the real-world datasets. All the models have node classification accuracy $\ge$ 0.90 for Noisy BA-House, $\ge$ 0.82 for Noisy BA-Community, $\ge$ 0.97 for Noisy Tree-Cycle, and $\ge$ 0.99 for Noisy Tree-Grid. Code and datasets will be available at \href{https://github.com/gisst/gisst}{https://github.com/gisst/gisst}.

\section{Additional explanation performance results for synthetic datasets}
\begin{table*}[h]
  \caption{Edge explanation accuracy results in synthetic datasets.}
  \label{tab:syn_edge_acc}
  \centering
  \addtolength{\tabcolsep}{-4pt}
  \begin{tabular}{lcccc}
    \toprule
    \textbf{Method}  & \textbf{Noisy BA-House}  & \textbf{ Noisy BA-Comm}  & \textbf{Noisy Tree-Cycle}  & \textbf{Noisy Tree-Grid} \\
    \midrule
    GAT              & 0.978              & 0.974                & 0.811                & 0.677              \\
    Grad             & 0.979              & 0.985                & 0.802                & 0.746              \\
    GNNExplainer     & 0.971              & \textbf{0.986}       & 0.759                & 0.728              \\
    GISST            & \textbf{0.980}     & 0.985                & \textbf{0.812}       & \textbf{0.754}     \\
    \bottomrule
  \end{tabular}
\end{table*}

\begin{table*}[h]
  \caption{Edge explanation recall results in synthetic datasets.}
  \label{tab:syn_edge_recall}
  \centering
  \addtolength{\tabcolsep}{-4pt}
  \begin{tabular}{lcccc}
    \toprule
    \textbf{Method}  & \textbf{Noisy BA-House}  & \textbf{ Noisy BA-Comm}  & \textbf{Noisy Tree-Cycle}  & \textbf{Noisy Tree-Grid} \\
    \midrule
    GAT              & 0.250              & 0.258                 & \textbf{0.571}       & 0.299             \\
    Grad             & 0.275              & 0.201                 & 0.377                & 0.359             \\
    GNNExplainer     & 0.217              & \textbf{0.322}        & 0.446                & \textbf{0.369}    \\
    GISST            & \textbf{0.288}     & 0.228                 & 0.382                & 0.365             \\
    \bottomrule
  \end{tabular}
\end{table*}

\begin{table*}[h]
  \caption{Node feature explanation accuracy results in synthetic datasets.}
  \label{tab:syn_edge_acc}
  \centering
  \addtolength{\tabcolsep}{-4pt}
  \begin{tabular}{lcccc}
    \toprule
    \textbf{Method}  & \textbf{Noisy BA-House}  & \textbf{ Noisy BA-Comm}  & \textbf{Noisy Tree-Cycle}  & \textbf{Noisy Tree-Grid} \\
    \midrule
    GAT              & NA                 & NA                   & NA                   & NA                 \\
    Grad             & \textbf{0.680}              & \textbf{0.680}                & 0.600                & 0.760              \\
    GNNExplainer     & 0.600              & 0.640                & 0.720                & 0.680              \\
    GISST            & 0.600              & 0.600                & \textbf{0.800}       & \textbf{0.880}     \\
    \bottomrule
  \end{tabular}
\end{table*}

\begin{table*}[h]
  \caption{Node feature explanation recall results in synthetic datasets.}
  \label{tab:syn_edge_recall}
  \centering
  \addtolength{\tabcolsep}{-4pt}
  \begin{tabular}{lcccc}
    \toprule
    \textbf{Method}  & \textbf{Noisy BA-House}  & \textbf{ Noisy BA-Comm}  & \textbf{Noisy Tree-Cycle}  & \textbf{Noisy Tree-Grid} \\
    \midrule
    GAT              & NA                      & NA                        & NA                   & NA                 \\
    Grad             & \textbf{0.800}          & \textbf{0.800}            & 0.750                & 0.850              \\
    GNNExplainer     & 0.750                   & 0.775                     & 0.825                & 0.800              \\
    GISST            & 0.750                   & 0.750                     & \textbf{0.875}       & \textbf{0.925}     \\
    \bottomrule
  \end{tabular}
\end{table*}

\end{document}